\documentclass[runningheads]{llncs}
\pdfoutput=1

\usepackage{graphicx}
\usepackage{amsmath}
\usepackage{acronym}
\usepackage[inline]{enumitem}
\usepackage{multirow}
\usepackage{graphicx}
\usepackage[dvipsnames]{xcolor}
\usepackage{bbm}
\usepackage{graphicx}
\usepackage{booktabs}
\usepackage[skip=0pt]{caption}
\usepackage{hyperref}
\usepackage{scrextend}
\usepackage[sectionbib,numbers,square,sort&compress]{natbib}

\usepackage[moderate,lists=normal,leading=normal]{savetrees}

\usepackage[para]{footmisc}

\AtBeginDocument{%
  \providecommand\BibTeX{{%
    \normalfont B\kern-0.5em{\scshape i\kern-0.25em b}\kern-0.8em\TeX}}}

\acrodef{ICR}{image-caption retrieval}
\acrodef{i2t}{image-to text}
\acrodef{t2i}{text-to-image}
\acrodef{mAP}{mean Average Precision}
\acrodef{DPR}{Dense Passage Retrieval}

\renewcommand{\vec}[1]{\mathbf{#1}}

\newcommand{\header}[1]{\vspace{1mm}\noindent\textbf{#1}.}
\newcommand{\subheader}[1]{\vspace{1mm}\indent\textbf{#1}.}

\newcommand{\AnalysisMethodLong}{counting contributing samples}
\newcommand{\AnalysisMethod}{COCOS}

\newcommand{\shrink}{\vspace*{-2mm}}

\allowdisplaybreaks

\begin{document}

\title{Do Lessons from Metric Learning Generalize\\ to Image-Caption Retrieval?}


\author{Maurits Bleeker \and Maarten de Rijke}
\institute{
	University of Amsterdam, Amsterdam, The Netherlands\\
	\email{\{m.j.r.bleeker, m.derijke\}@uva.nl}}

\authorrunning{M. Bleeker \and M. de Rijke}

\maketitle

\begin{abstract}
	The triplet loss with semi-hard negatives has become the de facto choice for \ac{ICR} methods that are optimized from scratch. 
	Recent progress in metric learning has given rise to new loss functions that outperform the triplet loss on tasks such as image retrieval and representation learning. 
	We ask whether these findings generalize to the setting of \ac{ICR} by comparing three loss functions on two \ac{ICR} methods.
	We answer this question negatively: the triplet loss with semi-hard negative mining still outperforms newly introduced loss functions from metric learning on the \ac{ICR} task.
	To gain a better understanding of these outcomes, we introduce an analysis method to compare loss functions by counting how many samples contribute to the gradient w.r.t.\ the query representation during optimization. 
	We find that loss functions that result in lower evaluation scores on the \ac{ICR} task, in general,  take too many (non-informative) samples into account when computing a gradient w.r.t.\ the query representation, which results in sub-optimal performance. 
	The triplet loss with semi-hard negatives is shown to outperform the other loss functions, as it only takes one (hard) negative into account when computing the gradient.
\end{abstract}


\section{Introduction}
\label{section:intro}

\shrink
Given a query item in one modality, \emph{cross-modal retrieval} is the task of retrieving similar items in another modality~\cite{zeng2020deep}. 
We focus on \acfi{ICR}~\citep{li2019visual, diao2021similarity, lee2018stacked,verma-2020-using}. 
For the \ac{ICR} task, given an image or a caption as a query, systems have to retrieve the positive (e.g., matching or similar) item(s) in the other modality.
Most \ac{ICR} methods work with a separate encoder for each modality to map the input data to a representation in a shared latent space~\citep{diao2021similarity, li2019visual, faghri2018vse++, jia2021scaling, lee2018stacked}. 
The encoders are optimized by using a contrastive-loss criterion, so as to enforce a high degree of similarity between  representations of matching items in the latent space.
For retrieval, a similarity score between a query and each candidate in a candidate set is computed to produce a ranking with the top-$k$ best matching items. 
A lot of recent work on \ac{ICR} relies on 
\begin{enumerate*}[label=(\arabic*)]
	\item pre-training on large amounts of data \cite{jia2021scaling, radford2021learning, li2020oscar}, and 
	\item more sophisticated (and data-hungry) model architectures \cite{messina2021transformer, faghri2018vse++, li2019visual, diao2021similarity, lee2018stacked, chen2020intriguing}.
\end{enumerate*}
However, pre-training on large-scale datasets is not always an option, either due to a lack of compute power, a lack of data, or both.  
Hence, it is important to continue to develop effective \ac{ICR} methods that only rely on a modest amount of data.

To learn the similarity between a query and candidate representations, most \ac{ICR} work relies on the standard Triplet loss with semi-hard negatives (Triplet SH)~\citep{faghri2018vse++,  li2019visual, messina2021transformer, diao2021similarity, lee2018stacked, chen2020intriguing, chen2020imram} or on the cross-entropy based NT-Xent~\citep{chen2020adaptive, jia2021scaling} loss.
In \emph{metric learning}, the focus is on loss functions that result in more accurate item representations (in terms of a given evaluation metric) that can distinguish between similar and dissimilar items in a low-dimensional latent space~\citep{musgrave2020metric}.
There has been important progress in metric learning, with the introduction of new loss functions that result in better evaluation scores on a specific (evaluation) task.  
For example SmoothAP~\cite{brown2020smooth}, it is a smooth approximation of the discrete evaluation metric Average Precision. By using SmoothAP, a retrieval method can be optimized with a discrete ranking evaluation metric and can handle multiple positive candidates simultaneously, which is not possible for the standard Triplet loss. 
Loss functions such as SmoothAP narrow the gap between the training setting and a discrete evaluation objective and thereby improve evaluation scores. 

\header{Research goal}
Most metric learning functions work with general representations of similar/dissimilar candidates and, in principle, there is no clear argument why obtained results on a specific task/method should not generalize to other tasks or methods.
Hence, \textit{can newly introduced metric learning approaches, that is, alternative loss functions, be used to increase the performance of \ac{ICR} methods}? 
We compare three loss function for the \ac{ICR} task: 
\begin{enumerate*}[label=(\arabic*)]
\item the Triplet loss \cite{kiros2014unifying}, including semi-hard negative mining, 
\item NT-Xent loss~\cite{chen2020simple},  and
\item SmoothAP~\cite{brown2020smooth}.
\end{enumerate*}
We expect SmoothAP to result in the highest performance based on the findings in in the context of image retrieval \citep{brown2020smooth} and in representation learning~\citep{varamesh2020self}.

\header{Main findings}
Following~\citep{musgrave2020metric}, we evaluate the three loss functions on fixed methods, with different datasets, and with a fixed training regime (i.e., training hyper-parameters) to verify which loss function uses the given training data as effectively as possible.
Surprisingly, the lessons from metric learning do not generalize to \ac{ICR}.
The Triplet loss with semi-hard negative mining still outperforms the other loss functions that we consider. 
The promising results obtained by SmoothAP and the NT-Xent loss in other fields do not generalize to the \ac{ICR} task. 

To get a better grasp of this unexpected outcome, we propose \emph{\AnalysisMethodLong{}} (\AnalysisMethod{}), a method for analyzing contrastive loss functions. 
The gradient w.r.t.\ the query for the Triplet loss, NT-Xent and SmoothAP can be formulated as a sum over the representations of the positive and negative candidates in the training batch. 
The main difference between the loss functions lies in the number of samples used when computing the gradient w.r.t.\ the query and how each sample is weighted.
We compare loss functions by counting how many samples contribute to the gradient w.r.t.\ the query representation at their convergence points. 
This yields an explanation of why one loss function outperforms another on the \ac{ICR} task.

\header{Main contributions}
\begin{enumerate*}[label=(\arabic*)]
	\item We experimentally compare three loss functions from the metric learning domain to determine if promising results from metric learning generalize to the \ac{ICR} task, and find that the Triplet loss semi-hard (SH) still results in the highest evaluation scores.
	\item We propose \AnalysisMethod{}, a way of analyzing contrastive loss functions, by defining a count that tells us how many candidates in the batch contribute to the gradient w.r.t.\ the query. On average, the best performing loss function takes at most one (semi-hard) negative sample into account when computing the gradient.
\end{enumerate*}

\section{Background and Related Work}
\label{sec:background}

\shrink
\textbf{Notation.}
We  follow the notation introduced in \citep{brown2020smooth,chen2020simple, varamesh2020self}.
We start with a multi-modal image-caption dataset $\mathcal{D} = \{(\textbf{x}_{I}^{i}, \textbf{x}_{C_1}^{i}$, \dots, $\textbf{x}_{C_k}^{i})^{i}, \dots \}_{i=1}^{N}$  that contains $N$ image-caption tuples. 
For each image $\textbf{x}_{I}^{i}$, we have $k$ matching/corresponding captions, $\textbf{x}_{C_1}^{i}, \dots, \textbf{x}_{C_k}^{i}$. 

In the \ac{ICR} task, either an image or a caption can function as a query. 
Given a query $\vec{q}$, the task is to rank all candidates in a candidate set $\Omega = \{\vec{v}_{i} \mid i=0, \dots, m\}$. 
A matching candidate is denote as $\vec{v}^{+}$ and a negative candidate(s) as $\vec{v}^{-}$.
For each query $\vec{q}$, we can split the candidate set $\Omega$ into two disjoint subsets: $ \vec{v}^{+} \in \mathcal{P}_{\vec{q}}$ (\textit{positive} candidate set) and $  \vec{v}^{-} \in\mathcal{N}_{\vec{q}}$ (\textit{negative} candidate set), where $\mathcal{ N}_{\vec{q}} = \{ \vec{v}^{-}  \mid   \vec{v}^{-} \in \Omega,   \vec{v}^{-}  \notin  \mathcal{P}_{\vec{q}} \}$. 
We assume a binary match between images and captions, they either match or they do not match. 

The set with similarity scores for each $\vec{v}_{i} \in \Omega$ w.r.t.\ query $\vec{q}$ is defined as: $\mathcal{S}_{\Omega}^{\vec{q}} = \{ s_{i} = \langle \frac{\vec{q}}{\| \vec{q}\|}  \frac{\vec{v}_{i}}{\| \vec{v}_{i }\|} \rangle, i=0,\ldots, m \}$.
We use cosine similarity as a similarity scoring function.  
$S_{\Omega}^{\vec{q}}$ consists of two disjoint subsets: $S_{\mathcal{P}}^{\vec{q}}$ and $S_{\mathcal{N}}^{\vec{q}}$.  $S_{\mathcal{P}}^{\vec{q}}$ contains the similarity scores for the positive candidates and $S_{\mathcal{N}}^{\vec{q}}$ the similarity scores for the negative candidates. 
During training, we randomly sample a batch $\mathcal{B}$ with image-caption pairs. Both the images and captions will functions as queries and candidates.

\header{Image-caption retrieval}
The \ac{ICR} task can be divided into \acfi{i2t} and \acfi{t2i} retrieval. 
We target specific~\ac{ICR} methods that are optimized for the \ac{ICR}-task only and satisfy three criteria: 
\begin{enumerate*}[label=(\arabic*)]
	\item The methods we use have solely been trained and evaluated on the same benchmark dataset;  
	\item the \ac{ICR} methods we use compute one  global representation for both  the image  and caption; and 
	\item the methods do not require additional supervision signals besides the contrastive loss for optimization.
\end{enumerate*}
Below we evaluate two \ac{ICR} methods with different loss functions: VSE++ \cite{ faghri2018vse++} and VSRN \cite{li2019visual}. 
In the online appendix of this work,\footnote{\label{online_appendix}\url{https://github.com/MauritsBleeker/ecir-2022-reproducibility-bleeker/blob/master/appendix}} we provide a detailed description of VSE++ and VSRN.

\subheader{VSE++} The best performing method of VSE++ uses a ResNet-152~\citep{he2016deep}) to compute a global image representation.
	The caption encoder is a single directed GRU-based \cite{cho2014properties} encoder. 
	\citet{faghri2018vse++} introduce the notion of mining semi hard-negative triplets for the \ac{ICR} task.
	By using the hardest negative in the batch for each positive pair (i.e. the negative candidate with the highest similarity score w.r.t. the query), their method outperforms state-of-the-art methods that do not apply this semi-hard negative mining. 
	
\subheader{VSRN} VSRN takes a set of pre-computed image region features as input. 
	A Graph Convolutional Network \cite{kipf2016semi} is used to enhance the relationships between each region vector.
	The sequence of region feature vectors is put through an RNN network to encode the global image representation.
	VSRN uses the same caption encoder and loss as \cite{faghri2018vse++}.

\subheader{Other methods}
Following VSE++ and VSRN, the SGRAF \cite{diao2021similarity}  and IMRAM \cite{chen2020imram} methods have been introduced. 
We do not use these two methods as they either do not outperform VSRN~\cite{chen2020imram} or rely on similar principles as VSRN~\cite{diao2021similarity}.
The main recent progress in \ac{ICR} has been characterized by a shift towards transformer-based~\cite{vaswani2017attention}  methods.  
To the best of our knowledge, TREN/TERAN \cite{messina2020fine, messina2021transformer} and VisualSparta \cite{luvisualsparta} are the only transformer-based \ac{ICR} methods that are solely optimized using MS-COCO  \cite{lin2014microsoft}  or Flickr30k \cite{young2014image}. We do not use transformer-based methods, as optimizing them does not scale well for a reproduciblity study with moderately sized datasets.
Methods such as OSCAR \cite{li2020oscar}, UNITER \cite{chen2020uniter}, Vilbert \cite{lu2019vilbert} and ViLT-B \cite{kim2021vilt} use additional data sources and/or loss functions for training.
They focus on a wide variety of tasks such as visual QA, image captioning, and image retrieval.

\header{Loss functions for \ac{ICR}}
In this section we introduce three  loss functions for \ac{ICR}.

\subheader{Triplet loss with semi hard-negative mining} The Triplet loss is commonly used as a loss function for \ac{ICR} methods \cite{faghri2018vse++, li2019visual, lee2018stacked, messina2021transformer, diao2021similarity, chen2020intriguing}. The \emph{Triplet loss with semi-hard negative mining} (Triplet loss SH), for a query $\vec{q}$ is defined as:
\begin{equation}
\mathcal{L}_\mathit{TripletSH}^{\vec{q}} = \max(\alpha - s^{+} +  s^{-}, 0),
\end{equation}
where $\alpha$ is a margin parameter, $s^{-} = \max(S_{\mathcal{N}}^{\vec{q}} )$ and   $s^{+} =  s_{0} \in S_{\mathcal{P}}^{\vec{q}}$. 
Here, $\mathcal{S}_{\mathcal{P}}^{\vec{q}}$ only contains one element per query. The Triplet loss SH over the entire training batch is defined as:
\begin{equation}
\mathcal{L}_\mathit{TripletSH}= \sum_{\vec{q} \in \mathcal{B}} 	\mathcal{L}_\mathit{TripletSH}^{\vec{q}}.
\end{equation}
Triplet loss SH performs a form of soft-negative mining per query by selecting the negative candidate with the highest similarity score w.r.t.\ the query, we also refer to this as the maximum violating query. For computational efficiency, this soft-negative mining is executed within the context of the training batch $\mathcal{B}$ and not over the entire training set.

As opposed to the definition above, another possibility is to take the Triplet-loss over all triplets in the batch $\mathcal{B}$. 
This is the definition of the standard \emph{Triplet-loss} \citep{kiros2014unifying}:
\begin{subequations}
	\begin{align}
	\mathcal{L}_\mathit{Triplet}^{\vec{q}} &= \sum_{ s^{-} \in S_{\mathcal{N}}^{q }}\max(\alpha - s^{+} +  s^{-}, 0)
	\\
	\mathcal{L}_\mathit{Triplet} &= \sum_{\vec{q} \in \mathcal{B}} 	\mathcal{L}_\mathit{Triplet}^{\vec{q}}.
	\end{align}
\end{subequations}

\subheader{NT-Xent loss} The \emph{NT-Xent loss} \cite{chen2020simple} is a loss function commonly used in the field of self-supervised representation learning \cite{oord2018representation, chen2020simple}. A similar function has also been proposed by \citet{zhang2018deep} in the context of \ac{ICR}.
The NT-Xent loss is defined as: 
\begin{equation} \label{eq:ntxent}
\mathcal{L}_\mathit{NT\mbox{-}Xent} =- \frac{1}{ |\mathcal{B}|} \sum_{\vec{q} \in \mathcal{B}} \log \frac{\exp({s^{+}/ \tau )}}{ \sum_{
		s_{i} \in \mathcal{S}_{\Omega}^{\vec{q}}
	}^{} \exp({s_{i}/ \tau ) }},
\end{equation}
where $\tau$ functions as a temperature parameter. As for the Triplet-loss formulation: $s^{+} =  s_{0} \in S_{\mathcal{P}}^{\vec{q}}$. 
The major difference between the Triplet-loss SH is that the NT-Xent loss takes the entire negative candidate set into account. 

\subheader{SmoothAP loss}
The Average Precision metric w.r.t.\ a query $\vec{q}$ and candidate set $\Omega$ is defined as:
\begin{equation}
\textstyle
\label{eq:ap}
AP_{\vec{q}} = 
\frac{1}{|\mathcal{S}_{\mathcal{P}}^{\vec{q}}|} 
\sum_{i \in \mathcal{S}_{\mathcal{P}}^{\vec{\vec{q}}}} 
\frac{\mathcal{R}(i, \mathcal{S}_{\mathcal{P}}^{\vec{q}})}
{\mathcal{R}(i, \mathcal{S}_{\Omega}^{\vec{q}})},
\end{equation} 
where $\mathcal{R}(i, \mathcal{S})$ is a function that returns the ranking of candidate $i \in \mathcal{S}$ in the candidate set:
\begin{equation}
\textstyle
\mathcal{R}(i, \mathcal{S}) = 1 + \sum_{j \in \mathcal{S}, i \neq j} \mathbbm{1} \{s_{i} - s_{j} < 0 \}.
\end{equation}
Let us introduce the $M \times M$ matrix $D$, where $D_{ij} = s_{i} - s_{j}$. By using the matrix $D$, Eq.~\ref{eq:ap} can be written as:
\begin{equation} 
\textstyle
AP_{\vec{\vec{q}}} = {}
\frac{1}{|\mathcal{S}_{\mathcal{P}}^{\vec{q}}|} \!\sum_{\substack{i\in\mathcal{S}_{\mathcal{P}}^{\vec{q}}}} \!
\frac{1+ \sum_{j \in \mathcal{S}_{\mathcal{P}} ,j\neq i} \mathbbm{1}\{D_{ij} > 0\}}
{1+ \sum_{j\in \mathcal{S}_{\mathcal{P}}^{\vec{q}} ,j\neq i} \mathbbm{1}\{D_{ij} > 0\} 
	+ \sum_{j\in \mathcal{S}_{\mathcal{N}}^{\vec{q}} } \mathbbm{1}\{D_{ij} > 0\}}.
\nonumber
\end{equation}
The indicator function $\mathbbm{1} \{\cdot\}$ is non-differentiable.  
To overcome this problem, the indicator function can be replaced by a sigmoid function:
\begin{equation} \label{eq:sigmoid}
\mathcal{G}(x; \tau) = \frac{1}{1 + e^{\frac{-x}{\tau}}}.
\end{equation}
By replacing the indicator function $\mathbbm{1} \{\cdot\}$  by $\mathcal{G}$,  the Average Precision metric can be approximated with a smooth function:
\begin{equation}  
\textstyle
\label{eq:smooth}
AP_{\vec{q}} \approx {}
\frac{1}{|\mathcal{S}_{\mathcal{P}}^{q}|} \sum_{i\in\mathcal{S}_{\mathcal{P}}^{\vec{q}}} 
\frac{1 + \sum_{j \in \mathcal{S}_{\mathcal{P}}^{\vec{q}} , j\neq i} \mathcal{G}(D_{ij}; \tau)}
{1 + \sum_{j\in \mathcal{S}_{\mathcal{P}}^{\vec{q}}, j\neq i} \mathcal{G}(D_{ij}; \tau) 
	+ \sum_{j\in \mathcal{S}_{\mathcal{N}}^{\vec{q}}} \mathcal{G}(D_{ij}; \tau)}.
\nonumber		
\end{equation}
This loss function is called \emph{SmoothAP} and has been introduced in the context of image retrieval~\citep{brown2020smooth}, following similar proposals in document retrieval and learning to rank~\citep{bruch-2019-analysis,bruch-2019-revisiting,oosterhuis-2018-differentiable,wang-2018-lambdaloss}.
The total loss over a batch $\mathcal{B}$ can then be formulated as follows: 
\begin{equation}
\textstyle
\mathcal{L}_{AP}  = \frac{1}{|\mathcal{B}|}\sum_{\vec{q} \in \mathcal{B}} (1 - AP_\vec{q}).
\end{equation}
In the online appendix,\footref{online_appendix} we provide an extended explanation of SmoothAP.

\section{Do Findings from Metric Learning Extend to \ac{ICR}?}

\shrink
In representation learning it was found that NT-Xent loss outperforms the Triplet loss and Triplet loss SH \cite{chen2020simple}. For both the image retrieval and representation learning task, results show that SmoothAP outperforms both the Triplet loss SH and the NT-Xent loss \cite{brown2020smooth, varamesh2020self}. 
We examine whether these findings generalize to \ac{ICR}.

\header{Experimental setup}
We focus on two benchmark datasets for the \ac{ICR} task: the Flickr30k~\cite{young2014image} and MS-COCO~\cite{lin2014microsoft} datasets.
Similar to \cite{faghri2018vse++, li2019visual}, we use the split provided by \citet{karpathy2015deep} for MS-COCO and the Flickr30k.
For  details of the specific  implementations of VSE++ \cite{faghri2018vse++}\footnote{\url{https://github.com/fartashf/vsepp}}  and \cite{li2019visual}\footnote{\url{https://github.com/KunpengLi1994/VSRN}} we refer to the papers and online implementations.
Each method is trained for 30 epochs with a batch size of 128. We start with a learning rate of 0.0002 and after 15 epochs we lower the learning rate to 0.00002. 

For  VSE++, we do not apply additional fine-tuning of the image encoder after 30 epochs. 
Our main goal is to have a fair comparison across methods, datasets, and loss functions, not to have the highest overall evaluation scores.
For VSE++, we use ResNet50 \cite{he2016deep} as image-encoder instead of ResNet152 \cite{he2016deep} or VGG \cite{simonyan2014very}. 
ResNet50 is faster to optimize and the performance differences between ResNet50 and ResNet152 are relatively small.

The  VSRN method comes with an additional caption decoder, to decode the original input caption from the latent image representation, this to add additional supervision to the optimization process. We remove the additional image-captioning module, so as to exclude performance gains on the retrieval tasks due to this extra supervision. 
In \cite{li2019visual}, the similarity score for a query candidate pair, during evaluation, is based on averaging the predicted similarity scores of (an ensemble of) two trained models. 
We only take the predicted relevance score of one model. 
The reason for this is that the evaluation score improvements are marginal when using the scores of two models (instead of one) but optimizing the methods takes twice as long. 
Therefore, our results are lower than the results published in \cite{li2019visual}. 
For all the remaining details, we refer to our repository.\footnote{\url{https://github.com/MauritsBleeker/ecir-2022-reproducibility-bleeker}}
When optimizing with SmoothAP, we take all the $k$ captions into account when sampling a batch, instead of one positive candidate. For this reason, we have to increase the amount of training epochs $k$ times as well to have a fair comparison.
For each loss function, we select the best performing hyper-parameter according to its original work. 

\header{Experiments}
We evaluate each loss function we described in Section~\ref{sec:background} given a dataset and method.  
For ease of reference, we refer to each individual evaluation with an experiment number (\#) (see Table~~\ref{table:recall}).
To reduce the variance in the results we run each experiment five times and report the average score and standard deviation.
Similar to \cite{faghri2018vse++, li2019visual}, we evaluate using Recall@k with $k  =  \{1,5,10\}$, for both the \acf{i2t} and \acf{t2i} task.
We also report the sum of all the recall scores (rsum) and the average recall value. 
For the \ac{i2t} task, we also report the mean average precision at $5$ (mAP@5) due to the fact we have $k$ positive captions per image query.

\begin{table*}[t]
		\caption{Evaluation scores for the Flickr30k and MS-COCO, for the VSE++ and VSRN.}
		\label{table:recall}
		\centering
		\setlength{\tabcolsep}{1.75pt}
\resizebox{\textwidth}{!}{%
	\begin{tabular}{@{} lcl @{~} cccc cccc cc@{}}
	\toprule
		&  &  & \multicolumn{4}{c}{i2t} & \multicolumn{4}{c}{t2i} &  &  \\ 
		\cmidrule(r){4-8}\cmidrule{9-12}
		{Loss function} & {\#} & {\begin{tabular}[c]{@{}c@{}}hyper\\ param\end{tabular}} & R@1 & R@5 & R@10 & {\begin{tabular}[c]{@{}c@{}}average\\ recall\end{tabular}} &  {mAP@5} & R@1 & R@5 & R@10 & {\begin{tabular}[c]{@{}c@{}}average\\ recall\end{tabular}} & rsum \\ 
	   \midrule
		&& \multicolumn{11}{c}{Flickr30k} \\ 
		\cmidrule(r){4-13}
		& & \multicolumn{11}{c}{VSE++} \\ 
		{Triplet loss} & {1.1} & {$\alpha=0.2$} &  30.8$\pm$.7  &  62.6$\pm$.3  &  74.1$\pm$.8  & { 55.9$\pm$.3 } & {0.41$\pm$.00}  &  23.4$\pm$.3 &  52.8$\pm$.1  &  65.7$\pm$.3 & { 47.3$\pm$.1} & 309.4$\pm$0.9 \\
		{Triplet loss SH} & {1.2} & {$\alpha=0.2$} &  \textbf{42.4}$\pm$.5  &  \textbf{71.2}$\pm$.7  &  \textbf{80.7}$\pm$.7  & { 64.8$\pm$.6 } & { 0.50$\pm$.01 } & \textbf{30.0}$\pm$.3  &  \textbf{59.0}$\pm$.2  &  \textbf{70.4}$\pm$.4  & { 53.1$\pm$.2 } &  \textbf{353.8}$\pm$1.6  \\
		{NT-Xent} & {1.3} & {$\tau=0.1$} &  37.5$\pm$.6  &  68.4$\pm$.6  &  77.8$\pm$.5 & { 61.2$\pm$.3 } & { 0.47$\pm$.00 } & 27.0$\pm$.3  &  57.3$\pm$.3 &  69.1$\pm$.2 & {51.1$\pm$.2}  &  337.1$\pm$1.3 \\
		{SmoothAP} & {1.4} & {$\tau=0.01$} &  \textbf{42.1}$\pm$.8  &  \textbf{70.8}$\pm$.6  &  \textbf{80.6}$\pm$.8  & { 64.5$\pm$.4 } & { 0.50$\pm$.00 } &  29.1$\pm$.3  &  58.1$\pm$.1  &  69.7$\pm$.2  & { 52.3$\pm$.2 } &   350.4$\pm$1.7 \\ 
		& & \multicolumn{11}{c}{VSRN} \\ 
		{Triplet loss} & {1.5} & {$\alpha=0.2$} &  56.4$\pm$.7  &  83.6$\pm$.6 &  90.1$\pm$.2  & { 76.7$\pm$.5 } & { 0.63$\pm$.01 } &  43.1$\pm$.3  &  74.4$\pm$.3  &  83.1$\pm$.4  & { 66.9$\pm$.3 } &   430.7$\pm$1.8  \\
		{Triplet loss SH} & {1.6} & {$\alpha=0.2$} &  \textbf{68.3}$\pm$1.3  &  \textbf{89.6}$\pm$.7  &  \textbf{94.0}$\pm$.5  & { 84.0$\pm$.5 } &  { 0.73$\pm$.01 } &  \textbf{51.2}$\pm$.9  &  \textbf{78.0}$\pm$.6  &  \textbf{85.6}$\pm$.5  & { 71.6$\pm$.6 } &  \textbf{466.6}$\pm$3.3  \\
		{NT-Xent} & {1.7} & {$\tau=0.1$} &  50.9 $\pm$.5 &  78.9$\pm$.7  &  86.6$\pm$.4  & { 72.2$\pm$.4} &  { 0.59$\pm$.00 } &  40.6$\pm$.6 &  71.9$\pm$.2 &  81.7$\pm$.3  & { 64.7$\pm$.2 } &  410.6$\pm$1.5  \\
		{SmoothAP} & {1.8} & {$\tau=0.01$} &  63.1$\pm$1.0  &  86.6$\pm$.8  &  92.4$\pm$.5  & {  80.7$\pm$.7  } &  { 0.69$\pm$.00 } &  45.8$\pm$.2  &   73.7$\pm$.3 &  82.3$\pm$.2  & { 67.3$\pm$.1 } &  444.0$\pm$2.1  \\ 
			\cmidrule(r){4-13}
		&& \multicolumn{11}{c}{MS-COCO}\textsl{} \\ 
			\cmidrule(r){4-13}
		& & \multicolumn{11}{c}{VSE++} \\ 
		{Triplet loss} & {2.1} & {$\alpha=0.2$} &  22.1$\pm$.5  &  48.2$\pm$.3  &  61.7$\pm$.3  & { 44.0$\pm$.3 } & { 0.30$\pm$.00 } & 15.4$\pm$.1  &  39.5$\pm$.1  &  53.2$\pm$.1  & { 36.0$\pm$.1 }  &  240.0$\pm$0.9  \\
		{Triplet loss SH} & {2.2} & {$\alpha=0.2$} &  \textbf{32.5}$\pm$.2  &  \textbf{61.6}$\pm$.3  &  \textbf{73.8}$\pm$.3  & { 56.0$\pm$.2 } &  { 0.41$\pm$.00 } &  \textbf{21.3}$\pm$.1  &  \textbf{48.1}$\pm$.1  &  \textbf{61.5}$\pm$.0  & { 43.6$\pm$.1 } &  \textbf{298.8}$\pm$0.8  \\
		{NT-Xent} & {2.3} & {$\tau=0.1$} &  25.8$\pm$.5  &  53.6$\pm$.5  &  66.1$\pm$.2  & { 48.5$\pm$.3 } &  { 0.34$\pm$.00 } &   18.0$\pm$.1  &  43.0$\pm$.1  &  56.6$\pm$.2 & { 39.2$\pm$.1} &  263.0$\pm$0.9  \\
		{SmoothAP} & {2.4} & {$\tau=0.01$} &  30.8$\pm$.3  &  60.3$\pm$.2  &  \textbf{73.6}$\pm$.5  & { 54.9$\pm$.3 } & { 0.40$\pm$.00 }  &  20.3$\pm$.2  &  46.5$\pm$.2  &  60.1$\pm$.2  & { 42.3$\pm$.2 } &  291.5$\pm$1.4  \\ 
		& & \multicolumn{11}{c}{VSRN} \\ 
		{Triplet loss} & {2.5} & {$\alpha=0.2$} &  42.9$\pm$.4  &  74.3$\pm$.3  &  84.9$\pm$.4  & { 67.4$\pm$.3 } &  { 0.52$\pm$.00 } &  33.5$\pm$.1  &  65.1$\pm$.1 &  77.1$\pm$.2  & { 58.6$\pm$.1 } &  377.8$\pm$1.2  \\
		{Triplet loss SH} & {2.6} & {$\alpha=0.2$} &  \textbf{48.9}$\pm$.6  &  \textbf{78.1}$\pm$.5  &  \textbf{87.4}$\pm$.2  & { 71.4$\pm$.4 } &  { 0.57$\pm$.01 } &  \textbf{37.8}$\pm$.5  &  \textbf{68.1}$\pm$.5  &  \textbf{78.9}$\pm$.3  & { 61.6$\pm$.4 } &  \textbf{399.0}$\pm$2.3  \\
		{NT-Xent} & {2.7} & {$\tau=0.1$} &  37.9$\pm$.4  &  69.2$\pm$.2  &  80.7$\pm$.3  & { 62.6$\pm$.1 } &  { 0.47$\pm$.00 } &  29.5$\pm$.1  &  61.0$\pm$.2  &  74.0$\pm$.2 & { 54.6$\pm$.1 } &  352.3$\pm$0.5  \\
		{SmoothAP} & {2.8} & {$\tau=0.01$} &  46.0$\pm$.6  &  76.1$\pm$.3  &  85.9$\pm$.3  & { 69.4$\pm$.3 } &  { 0.54$\pm$.00 } &  33.8$\pm$.3  &  64.1$\pm$.1  &  76.0$\pm$.2 & { 58.0$\pm$.2 } &  382.0$\pm$1.1  \\ 
		\bottomrule
	\end{tabular}
}
\end{table*}

\subheader{Results}
Based on the scores reported in Table \ref{table:recall}, we have the following observations:
\begin{enumerate}[label=(\arabic*),nosep]
	\item Given a  fixed method and default hyper-parameters for each loss function, the Triplet loss SH results in the best evaluation scores, regardless of dataset, method or task. 
	\item Similar to \citep{faghri2018vse++}, we find that the Triplet loss SH consistently outperforms the general Triplet loss, which takes all the negative triplets in the batch into account that violate the margin constraint.
	\item The NT-Xent loss consistently underperforms compared to the Triplet loss SH. This is in contrast with findings in \cite{chen2020simple}, where the NT-Xent loss results in better down-stream evaluation performance on a (augmented image-to-image) representation learning task than the Triplet loss SH. 
	Although the \ac{ICR} task has different (input) data modalities, the underlying learning object is the same for \ac{ICR} and augmented image-to-image representation learning (i.e., contrasting positive and negative pairs). 
	\item Only for the VSE++ method on the \ac{i2t} task, SmoothAP performs similar to the Triplet-loss SH.
	\item SmoothAP does not outperform the Triplet loss SH. This is in contrast with the findings in \cite{brown2020smooth}, where SmoothAP does outperform Triplet-loss SH and other metric learning functions.
	\item The method with the best Recall@k score also has the highest mAP@k score. 
\end{enumerate} 

\subheader{Upshot}
Based on our observations concerning Table~\ref{table:recall}, we conclude the following:
\begin{enumerate*}[label=(\arabic*)]
	\item The Triplet loss SH should still be the \emph{de facto} choice for optimizing \ac{ICR} methods. 
	\item The promising results from the representation learning field that were obtained by using the NT-Xent loss \cite{chen2020simple}, do not generalize to the \ac{ICR} task.  
	\item Optimizing an \ac{ICR} method with a smooth approximation of a ranking metric (SmoothAP) does not result in better Recall@k scores. 
	\item Optimizing an \ac{ICR} method by using a pair-wise distance loss between the positive triplet and a semi-hard negative triplet still yields the best evaluation performance. For both methods VSE++ and VSRN, \ac{i2t} and \ac{t2i}  and for both datasets.
\end{enumerate*}


\shrink
\section{A Method for Analyzing the Behavior of Loss Functions}
\label{sec:grad_count}

\shrink
Next, we propose a method for analyzing the behavior of loss functions for \ac{ICR}.
The purpose is to compare loss functions, and explain the difference in performance.
If we compare the gradient w.r.t. $\vec{q}$ for the Triplet loss and the Triplet loss SH, the only difference is the number of triplets that the two loss functions take into account. 
If two models are optimized in exactly the same manner, except one model uses the Triplet loss and the other uses Triplet loss SH, the difference in performance can only be explained by the fact that the Triplet loss takes all violating triplets into account. 
This means that the number of triplets (i.e., candidates) that contribute to the gradient directly relates to the evaluation performance of the model. 
The same reasoning applies for the NT-Xent and the SmoothAP loss. 
For example, the gradient w.r.t.\ $\vec{q}$ for the NT-Xent loss also has the form $\vec{v}^{+} - \vec{v}^{-}$. The major difference between the two functions is that, for the negative candidate the NT-Xent loss computes a weighted sum over all negative to compute a representation of  $\vec{v}^{-}$. Therefore, the difference in evaluation performance between the Triplet loss SH and NT-Xent can only be explained by this weighted sum over all negatives. This  sum can be turned into a count of negatives, i.e., how many negative approximately contribute to this weighted sum, which can be related to the other losses. By counting the number of candidates that contribute to the gradient, we aim to get a better understanding of why a certain loss function performs better than others.
The method we propose is called \emph{\AnalysisMethodLong{}}~(\AnalysisMethod).

First, we provide the form of the derivative of each loss function w.r.t.\ query $\vec{q}$. For each loss function the derivative is a sum over $\vec{v}^{+} - \vec{v}^{-}$. Loss functions may weight the positive and negative candidate(s) differently, and the number of candidates or triplets that are weighted may differ across loss functions. 

\header{Triplet loss and Triplet loss SH}\label{par:gradtriplet}
The gradient w.r.t.\ $\vec{q}$ for the Triplet loss SH, $\mathcal{L}_\mathit{TripletSH}^{\vec{q}}$ is the difference between the representation of the positive and negative candidate:
\begin{subequations}
\begin{align}
\textstyle
\frac{\partial \mathcal{L}_\mathit{TripletSH}^{\vec{q}}}{\partial \vec{q}} &= 
\left\{
\begin{array}{ll}
\vec{v}^{+} - \vec{v}^{-}, & \textrm{if } s^{+} -  s^{-} < \alpha \\
0, & \textrm{otherwise}.
\end{array}
\right. \label{eq:triplet_grad}\\ 
\textstyle
\frac{\partial \mathcal{L}_\mathit{Triplet}^{\vec{q}}}{\partial \vec{q}} 
  & \textstyle = \sum_{\vec{v}^{-} \in \mathcal{N}_{\vec{q}}}  \mathbbm{1} \{s^{+} - s^{-} < \alpha \} \left( 	\vec{v}^{+} - \vec{v}^{-} \right). \label{eq:triplet_grad_2}
\end{align}
\end{subequations}
The gradient of Triplet loss $\mathcal{L}_\mathit{Triplet}^{\vec{q}}$ (Eq.~\ref{eq:triplet_grad_2}) w.r.t.\ $\vec{q}$ has a similar form. However, there the gradient is a sum over all triplets that violate $s^{+} +  s^{-} < \alpha$, and not only the maximum violating one.
Based on Eq.~\ref{eq:triplet_grad} we can see that a query $\vec{q}$ only has a non-zero gradient when $s^{+} - s^{-} < \alpha$. 
If this is the case, the gradient always has the form $\vec{v^{+}} - \vec{v^{-}}$, and this value is  independent of the magnitude $s^{+} - s^{-}$.
For this reason, given a batch $\mathcal{B}$, the number of queries $\vec{q}$ that have a non-zero gradient is defined by:
\begin{equation}
\textstyle
C^{\mathcal{B}}_{TripletSH} = \sum_{\vec{q} \in \mathcal{B}}    \mathbbm{1} \{ s^{+} - s^{-} < \alpha \},
\end{equation}
where $s^{+} = s^{0} \in \mathcal{S}_{\mathcal{P}}^{\vec{q}}$ and $s^{-} = \max \left( \mathcal{S}_{\mathcal{N}}^{\vec{q}}  \right)$.  
We define $ C^{\mathcal{B}}_{TripletSH} $ to be \textit{the number of queries $\vec{q}$ that have a non-zero gradient given batch $\mathcal{B}$.}

As the Triplet loss takes all the triplets into account that violate the distance margin $\alpha$, we can count three things: 
\begin{enumerate*}[label=(\arabic*)]
	\item Per query $\vec{q}$, we can count how many triplets $\vec{v}^{+} - \vec{v}^{-}$ contribute to the gradient of $\vec{q}$. We define this as $C_{Triplet}^{\vec{q}} = \sum_{s^{-} \in \mathcal{S}_{\mathcal{N}}^{\vec{q}}}  \mathbbm{1} \{ s^{+} - s^{-} < \alpha \}$. 
	\item Given the batch $\mathcal{B}$, we can count how many triplets contribute to the gradient over the entire training batch $\mathcal{B}$. We define this number as  $C_{Triplet}^{\mathcal{B}} = \sum_{\vec{q} \in \mathcal{B}} C_{Triplet}^{\vec{q}} $.
	\item Given the entire batch $\mathcal{B}$, we can count how many queries have a gradient value of zero (i.e., no violating triplets).  This number is  $C_{Triplet}^{0}  = \sum_{\vec{q} \in \mathcal{B}}  \mathbbm{1} \{  C_{Triplet}^{\vec{q}} = 0 \}$.
\end{enumerate*}

\header{NT-Xent loss}
The gradient w.r.t.\ $\vec{q}$ for the NT-Xent loss is defined as \cite{chen2020simple}:
\begin{equation}
\textstyle
\frac{\partial \mathcal{L}_\mathit{NT\mbox{-}Xent}^{\vec{q}}}{\partial \vec{q}} = 
\left( 1 - \frac{\exp(s^{+}/\tau)}{Z(\vec{q})} \right) \tau^{-1} \vec{v}^{+} -  \sum_{s^{-} \in S_{\mathcal{N}}^{q }} \left( \frac{\exp(s^{-}/\tau)}{Z(\vec{q})} \right)  \tau^{-1} \vec{v}^{-},
\label{eq:ntxent_grad}
\end{equation}
where $Z(\vec{q}) =   \sum_{s_{i} \in \mathcal{S}_{\Omega}^{\vec{q}}}^{} \exp({s_{i}/ \tau ) }$, a normalization constant depending on $\vec{q}$. 
The gradient w.r.t.\ $\vec{q}$ is the weighted difference of the positive  candidate $\vec{v}^{+}$ and the weighted sum over all the negative candidates.
The weight for each candidate is based on the similarity with the query, normalized by the sum of the similarities of all candidates. 
In contrast, for the Triplet-loss (Eq.~\ref{eq:triplet_grad_2}) all candidates are weighted equally when they violate the margin constraint. 
The NT-Xent loss performs a natural form of (hard) negative weighting~\citep{chen2020simple}. 
The more similar a negative sample is to the query, the higher the weight of this negative in the gradient computation. 
In principle, all the negatives and the positive candidate contribute to the gradient w.r.t.\ $\vec{q}$. 
In practice, most similarity scores $s^{-} \in \mathcal{S}_{\mathcal{N}}^{\vec{q}}$ have a low value; so the weight of this negative candidate in the gradient computation will be close to 0.

To count the  number of negative candidates that contribute to the gradient, we define a threshold value $\epsilon$.  
If the weight of a negative candidate $\vec{v}^{-}$ is below $\epsilon$, we assume that its contribution is negligible.
All candidate vectors are normalized. 
Hence, there is no additional weighting effect by the magnitude of the vector. 
For the NT-Xent loss we define three terms: 
$C_{NTXent}^{\vec{qv}^{-}} $, $W_{NTXent}^{\vec{qv}^{-}}$ and $W_{NT-Xent}^{\vec{qv}^{+}}$:
\begin{enumerate*}[label=(\arabic*)]
	\item Given a query $\vec{q}$, $C_{NT-Xent}^{\vec{qv}^{-}}$ is the number of negative candidates $\vec{v}^{-}$ that contribute to the gradient w.r.t.\ $\vec{q}$:
	$C_{NT-Xent}^{\vec{qv}^{-}}= \sum_{s^{-} \in  \mathcal{S}_{\mathcal{N}}^{\vec{q}}} \mathbbm{1} \{ \exp(s^{-}/\tau) Z(\vec{q})^{-1}  > \epsilon \}.$
	\item Given $C_{NT-Xent}^{\vec{qv}^{-}}$, we compute the sum of the weight values of the contributing negative canidates $\vec{v}^{-}$ as $W_{NT-Xent}^{\vec{qv}^{-}} =  \sum_{s^{-} \in  \mathcal{S}_{\mathcal{N}}^{\vec{q}}} \mathbbm{1} \{ \exp(s^{-}/\tau)Z(\vec{q})^{-1}  > \epsilon \}  \exp(s^{-}/\tau)Z(\vec{q})^{-1}$.
	\item We define
$W_{NT-Xent}^{\vec{qv}^{+}} = {}$ $\frac{1}{N} \sum_{\vec{q} \in \mathcal{B}}  ( 1 -  \exp(s^{+}/\tau)Z(\vec{q})^{-1} )$, as the mean weight value of the positive candidates in batch $\mathcal{B}$.
\end{enumerate*}

We define the two extra terms, $W_{NT-Xent}^{\vec{qv}^{-}}$ and $W_{NT-Xent}^{\vec{qv}^{+}}$ because for the NT-Xent function we have to count the candidates with a weight value above the threshold $\epsilon$. 
This count on its own does not provide a good picture of the contribution of these candidates to the gradient. 
Therefore, we compute a mean value of those weight values as well, to provide insight into the number of the samples on which the gradient w.r.t.\ $\vec{q}$ is based.

\header{SmoothAP loss}
A full derivation of the gradient of SmoothAP w.r.t.\ $\vec{q}$ is provided with the implementation of our methods.\footref{online_appendix} We introduce $sim(D_{ij})$, the derivative of~\eqref{eq:sigmoid}:
\begin{equation}\label{eq:smoothder}
\textstyle
	\begin{split}
	\textstyle
	\frac{\partial AP_{\vec{q}}}{\partial \vec{q}} & = {}
	\textstyle
	\frac{1}{|\mathcal{S}_{\mathcal{P}}^{\vec{q}}|} \sum_{\substack{i\in\mathcal{S}_{\mathcal{P}}^{\vec{q}}}} \mathcal{R}\left(i, \mathcal{S}_{\Omega}^{\vec{q}}\right)^{-2} \left( 
	\mathcal{R}\left(i, \mathcal{S}_{\mathcal{P}}^{\vec{q}}\right)
	\left( \sum_{j\in \mathcal{S}_{\mathcal{N}}^{\vec{q}}} sim(D_{ij})  ( \vec{v}_{i} - \vec{v}_{j}) \right) \right.
	-  \\	
	&\textstyle	\mbox{}\hspace*{36mm} \left.\left( \mathcal{R}\left(i, \mathcal{S}_{\mathcal{N}}^{\vec{q}}\right) - 1 \right) 
	\left( \sum_{j\in \mathcal{S}_{\mathcal{P}}^{\vec{q}}, j\neq i}  sim(D_{ij})  ( \vec{v}_{i} - \vec{v}_{j}) \right) \right).
		\end{split}
\end{equation}
Given Eq.~\ref{eq:smoothder}, it is less trivial to infer what the update w.r.t.\ $\vec{q}$ looks like in terms of positive candidates $\vec{v}_{i}$ and negative candidates $\vec{v}_{j}$. However, we can derive the following two properties:
\begin{enumerate*}[label=(\arabic*)]
	\item The lower a positive candidate $\vec{v}_{i}$ is in the total ranking, the less this candidate is taken into account for the gradient computation w.r.t.\ $\vec{q}$, due the inverse quadratic term $\mathcal{R}\left(i, \mathcal{S}_{\Omega}^{q}\right)^{-2}$. This is in line with optimizing the AP as a metric; positive candidates that are ranked low contribute less to the total AP score, and therefore are less important to optimize.
	\item Each triplet  $\vec{v}_{i} - \vec{v}_{j}$ is weighted according to their difference in similarity score $D_{ij}$. If their difference in similarity score w.r.t.\ query $\vec{q}$ is relatively small (i.e., $D_{ij}$ is close to zero), $sim(D_{ij})$ will have a high value due to the fact that $sim(D_{ij})$  is the derivative of the sigmoid function. Therefore,  $sim(D_{ij})$ indicates how close the similarity score (with the query) of candidate $\vec{v}_{i}$  is compared to the similarity score of $\vec{v}_{j}$  This is in line with the SmoothAP loss because we use a sigmoid to approximate the step-function; only triplets of candidates that have a similar similarity score will contribute to the gradient. 
\end{enumerate*} 

We define a threshold value $\epsilon$ again. 
If the value of $sim(D_{ij})$ is lower than the threshold value, we consider the contribution of this triplet to be negligible. We have to take into account that all triplets are also weighted by $\mathcal{R}(i, \mathcal{S}_{\Omega}^{\vec{q}})^{-2}$, which is always lower than or equal to 1. 
We can define $C_{Smooth}^{\vec{q}}$, which \textit{is the number of triplets $\vec{v}^{+} - \vec{v}^{-}$ that contribute to the gradient w.r.t. $\vec{q}$}, for SmoothAP as follows:
\begin{equation}
\textstyle
C_{Smooth}^{\vec{q}} = \frac{1}{|\mathcal{S}_{\mathcal{P}}^{\vec{q}}|} \sum_{i \in \mathcal{S}_{\mathcal{P}}^{\vec{q}}} \left(\sum_{j \in \mathcal{S}_{\mathcal{N}}^{\vec{q}}}  \mathbbm{1} \left\{ \frac{sim(D_{ij})}{\mathcal{R}(i, \mathcal{S}_{\Omega}^{\vec{q}})^{2}} > \epsilon\right\} 
+ \sum_{j \in \mathcal{S}_{\mathcal{P}}^{\vec{q}}, j \neq i}  \mathbbm{1} \left\{ \frac{sim(D_{ij})}{\mathcal{R}(i, \mathcal{S}_{\Omega}^{\vec{q}})^{2}} > \epsilon\right\} \right).
\end{equation}
Similar to \citep{brown2020smooth}, we use $sim(D_{ij})$ in combination with a threshold value $\epsilon$ to indicate which samples have a non-zero gradient in the training batch. 
We ignore the terms $\mathcal{R}\left(i, \mathcal{S}_{\mathcal{P}}^{\vec{q}} \right)$ and  $1 - \mathcal{R}\left(i, \mathcal{S}_{\mathcal{N}}^{\vec{q}} \right)$ for this gradient computation. 
We also count all queries $\vec{q}$ within batch $\mathcal{B}$ that do not have a gradient value. 
We define this number as $C_{Smooth}^{0} = \sum_{\vec{q} \in \mathcal{B}} \mathbbm{1} \{ C_{Smooth}^{\vec{q}} = 0 \}$. 
This completes the definition of \AnalysisMethod{}: for every loss function that we consider, it counts the number of candidates that contribute to the gradient w.r.t.\ $\vec{q}$.


\shrink\shrink
\section{Analyzing the Behavior of Loss Functions for \ac{ICR}}

\shrink
\textbf{Experimental setup.}
To use \AnalysisMethod~, we introduce the following experimental setup.
For each loss function, we take the checkpoint of one of the five optimized models. We refer to this checkpoint as \textit{the optimal convergence point} for this loss function.
This is not the point with the lowest loss value, but the model checkpoint that results in the highest evaluation scores on the validation set.
We freeze all model parameters and do not apply dropout. We iterate over the entire training set by sampling random batches $\mathcal{B}$ (with batch size $|\mathcal{B}| = 128$, similar to the training set-up). For each batch we compute the \AnalysisMethod{} and weight values defined in Section~\ref{sec:grad_count}. 
We report the mean value and standard deviation over the entire training set for both VSE++ and VSRN, for both datasets and for each loss function. 
The only hyper-parameter for this experiment is $\epsilon$.
We use $\epsilon=0.01$ for both the NT-Xent and SmoothAP loss.

\header{Experimental outcomes}
\begin{table*}[t]
	\caption{\AnalysisMethod ~w.r.t. query $\vec{q}$, for the Triplet loss and the Triplet loss SH.}
	\label{tab:gradcount_triplet}
	\centering
	\setlength{\tabcolsep}{3pt}
	\resizebox{\textwidth}{!}{
	\begin{tabular}{lll c ccc ccc}
	\toprule
		&  &  &  & \multicolumn{3}{c}{i2t} & \multicolumn{3}{c}{t2i} \\ 
		\cmidrule(r){5-7}\cmidrule{8-10}
		&  &  & \# & $C_{}^{\vec{q}}$ & $C_{}^{\mathcal{B}}$ & $C_{}^{0}$ & $C_{}^{\vec{q}}$ & $C_{}^{\mathcal{B}}$ & $C_{}^{0}$ \\ 
		\midrule
		\multirow{4}{*}{Flickr30k} & \multirow{2}{*}{VSE++} & \multicolumn{1}{l}{Triplet loss} & 1.1 & 6.79$\pm$0.83 & 768.92$\pm$96.87 & 14.78$\pm$3.52 & 6.11$\pm$0.75 & 774.67$\pm$98.05 & 1.14$\pm$1.22 \\
		\multicolumn{1}{c}{} &  & \multicolumn{1}{l}{Triplet loss SH} & 1.2 & 1$\pm$0.0 & 98.74$\pm$4.83 & 29.23$\pm$4.81 & 1$\pm$0.0 & 98.22$\pm$4.66 & 29.75$\pm$4.62 \\ 
		\cmidrule{2-10} 
		\multicolumn{1}{c}{} & \multirow{2}{*}{VSRN} & \multicolumn{1}{l}{Triplet loss} & 1.5 & 1.39$\pm$0.12 & 60.96$\pm$10.30 & 84.29$\pm$5.80 & 1.28$\pm$0.10 & 61.21$\pm$10.01 & 80.15$\pm$6.35 \\
		\multicolumn{1}{c}{} &  & \multicolumn{1}{l}{Triplet loss SH} & 1.6 & 1$\pm$0.0 & 45.59$\pm$5.93 & 82.39$\pm$5.92 & 1$\pm$0.0 & 44.98$\pm$5.70 & 82.99$\pm$5.70 \\ 
		\midrule
		\multirow{4}{*}{MS-COCO} & \multirow{2}{*}{VSE++} & \multicolumn{1}{l}{Triplet loss} & 2.1 & 3.51$\pm$0.49 & 353.82$\pm$52.71 & 27.09$\pm$4.60 & 2.94$\pm$0.36 & 341.64$\pm$50.80 & 12.24$\pm$4.92 \\
		&  & \multicolumn{1}{l}{Triplet loss SH} & 2.2 & 1 $\pm$0.0 & 88.17$\pm$5.25 & 39.82$\pm$5.24 & 1$\pm$0.0 & 87.24$\pm$5.34 & 40.75$\pm$5.33 \\ 
		\cmidrule{2-10} 
		& \multirow{2}{*}{VSRN} & \multicolumn{1}{l}{Triplet loss} & 2.5 & 1.21$\pm$0.13 & 29.88$\pm$7.46 & 103.33$\pm$5.22 & 1.15$\pm$0.10 & 30.25$\pm$7.49 & 101.70$\pm$5.58 \\
		&  & \multicolumn{1}{l}{Triplet loss SH} & 2.6 & 1$\pm$0.0 & 33.24$\pm$5.39 & 94.73$\pm$5.45 & 1$\pm$0.0 & 32.90$\pm$5.35 & 95.08$\pm$5.4 \\ 
		\bottomrule
	\end{tabular}
}
\end{table*}
For each of the loss functions that we consider, we analyze its performance using \AnalysisMethod{}.

\subheader{Triplet loss} Our goal is not to show that the Triplet loss SH outperforms the Triplet loss, which has already been shown~\cite{faghri2018vse++}, but to explain this behavior based on \AnalysisMethod{} w.r.t.\ $\vec{q}$ and also relate this to the NT-Xent and SmoothAP loss.

Based on Table~\ref{table:recall} (row 1.1/1.2 and 1.5/1.6, row 2.1/2.2 and 2.5/2.6) it is clear that the Triplet loss SH always outperforms the general Triplet loss with a large margin.
If we look at Table~\ref{tab:gradcount_triplet}, row 1.1/1.2 and 2.1/2.2, respectively, there is a clear relation between $\mathcal{C}^{\vec{q}}$ and the final evaluation score for the VSE++ model for both sub-tasks \ac{i2t} and \ac{t2i} (Table~\ref{table:recall}).
$\mathcal{C}_{Triplet}^{\vec{q}} $ and $\mathcal{C}_{Triplet}^{\mathcal{B}} $ are both much greater than $\mathcal{C}_{TripletSH}^{\vec{q}} $ and $\mathcal{C}_{TripletSH}^{\mathcal{B}}$, for both dataset and both the the \ac{i2t} and \ac{t2i} task. 
When multiple negatives with small margin violation are combined into a gradient, the gradient is dominated by easy or non-informative negative samples, which results in convergence of the model into a sub-optimal point~\citep{faghri2018vse++}.  
Clearly, the loss function with the lowest evaluation score takes into account the most negatives when computing the gradient w.r.t.\ $\vec{q}$.
Based on \citep{faghri2018vse++} and the \AnalysisMethod{} results in Table~\ref{tab:gradcount_triplet} we conclude that, at the optimal convergence point, the Triplet loss takes too many negatives into account (i.e., too many triplets still violate the margin constraint), leading to lower evaluation scores. 

For VSRN the relation between $\mathcal{C}_{Triplet}^{\vec{q}} $, $\mathcal{C}_{TripletSH}^{\vec{q}} $ and the final evaluation score is less clear. 
If we look at Table~\ref{tab:gradcount_triplet}, row 1.5/1.6 and 2.5/2.6, respectively, we see that $C_{Triplet}^{\vec{q}} \approx C_{TripletSH}^{\vec{q}}  =  1$. 
This means that at the optimal convergence point, for VSRN, the Triplet loss and the Triplet loss SH (approximately) are a similar to each other and both functions only take one negative triplet into account when computing the gradient w.r.t.\ $\vec{q}$. 
Thus, both functions should result in approximately the same gradient value while the Triplet loss SH still outperforms the Triplet loss with a large margin.
This can be explained as  follows: At the start of training, for each query $\vec{q}$ (almost) all triplets violate the margin constraint (because all candidate representations are random). 
Therefore, the gradient(s) computation w.r.t.\ $\vec{q}$ for the Triplet loss is based on all triplets in the batch and therefore this gradient is dominated by a majority of non-informative samples in the beginning of the training, which leads to convergence at a sub-optimal point. 

\begin{table*}[t]
	\caption{\AnalysisMethod~w.r.t.\ query $\vec{q}$, for the NT-Xent loss \cite{chen2020simple}.}
	\label{tab:gradcount_ntxent}
	\setlength{\tabcolsep}{3pt}
	\centering
	\resizebox{\textwidth}{!}{
\begin{tabular}{llc ccc ccc}
\toprule
	&  &  & \multicolumn{3}{c}{i2t} & \multicolumn{3}{c}{t2i} \\ 
	\cmidrule(r){4-6}\cmidrule{7-9}
	&  & \# & $C_{NT-Xent}^{\vec{q}\vec{v}^{-}}$ & $W_{NT-Xent}^{\vec{q}\vec{v}^{-}}$ & $W_{NT-Xent}^{\vec{q}\vec{v}^{+}}$ & $C_{NT-Xent}^{\vec{q}\vec{v}^{-}}$ & $W_{NT-Xent}^{\vec{q}\vec{v}^{-}} $ & $W_{NT-Xent}^{\vec{q}\vec{v}^{+}} $ \\ 
	\midrule
	\multirow{2}{*}{Flickr30k} & VSE++ & 1.3 & 9.88$\pm$0.51 & 0.42$\pm$0.02 & 0.56$\pm$0.02 & 9.65$\pm$0.51 & 0.42$\pm$0.02 & 0.56$\pm$0.02 \\
	& VSRN & 1.7 & 2.45$\pm$0.23 & 0.13$\pm$0.02 & 0.20$\pm$0.02 & 2.46$\pm$0.23 & 0.13$\pm$0.02 & 0.20$\pm$0.02 \\ 
	\midrule
	\multirow{2}{*}{MS-COCO} & VSE++ & 2.3 & 5.59$\pm$0.40 & 0.36$\pm$0.02 & 0.46$\pm$0.02 & 5.33$\pm$0.38 & 0.36$\pm$0.02 & 0.46$\pm$0.02 \\
	& VSRN & 2.7 & 1.10$\pm$0.14 & 0.10$\pm$0.02 & 0.14$\pm$0.02 & 1.11$\pm$0.14 & 0.09$\pm$0.02 & 0.14$\pm$0.02 \\ 
\bottomrule
\end{tabular}
}
\end{table*}

\subheader{NT-Xent} Based on Table~\ref{tab:gradcount_ntxent}, we can see that $C_{NT-Xent}^{\vec{qv}^{-}}$ is higher than 1 for both VSE++ and VSRN, for \ac{i2t} and \ac{t2i}, on both datasets. 
If we relate the evaluation performances of the NT-Xent loss (row 1.3, 1.7, 2.3, 2.7) to the Triplet loss SH (row 1.2, 1.6, 2.2, 2.6)  in Table~\ref{table:recall}, we can see that the Triplet loss SH consistently outperforms the NT-Xent loss, regardless of the method, dataset or sub-task. We therefore can conclude that taking only the most violating negative into account when computing the gradient w.r.t. $\vec{q}$ results in better evaluation performances than computing a weighted sum over all negative candidates. We can apply the same reasoning used to explain the performance difference between the Triplet loss and Triplet loss SH. The gradient w.r.t.\ $\vec{q}$ for the NT-Xent is dominated by too many non-informative negatives, which have a weight value bigger than $\epsilon$.

Looking at Table~\ref{table:recall}, we see that NT-Xent loss outperforms the Triplet loss for the VSE++ method (1.3/1.1 and 2.3/2.1), while taking more negative samples into account when computing the  gradient (based on our definition of~\AnalysisMethod). 
This in contrast with the previous observation for the Triplet loss of the more (non-informative) samples a loss function takes into account when computing the gradient w.r.t. $\vec{q}$, the lower the evaluation score. Solely counting the number of negative examples that contribute to the gradient does not the provide the full picture  for the NT-Xent loss; the weight value of each individual sample (including the positive) plays a more important role than initially was assumed.
We have tried different values for $\epsilon$, with little impact.

\begin{table*}[t]
\caption{\AnalysisMethod~w.r.t.\ query $\vec{q}$, for the  SmoothAP \cite{brown2020smooth} loss.}
\label{tab:gradcount_smooth}
	\centering
	\setlength{\tabcolsep}{3pt}
\begin{tabular}{llc cc cc}
\toprule
	&  &  & \multicolumn{2}{c}{i2t} & \multicolumn{2}{c}{t2i} \\ 
	\cmidrule(r){4-5}\cmidrule{6-7}
	& \multicolumn{1}{l}{} & \# & $C_{SmoothAP}^{\vec{q}}$ & $C_{SmoothAP}^{0}$ & $C_{SmoothAP}^{\vec{q}}$ & $C_{SmoothAP}^{0}$ \\ 
	\midrule
	\multirow{2}{*}{Flickr30k} & \multicolumn{1}{l}{VSE++} & 1.4 & 1.27$\pm$0.06 & 2.15$\pm$1.51 & 1.47$\pm$0.83 & 636.72$\pm$18.72 \\
	& \multicolumn{1}{l}{VSRN} & 1.8  & 2.33$\pm$0.07 & 0.00$\pm$0.00 & 1.62$\pm$0.95 & 636.49$\pm$18.65 \\ 
	\midrule
	\multirow{2}{*}{MS-COCO} & \multicolumn{1}{l}{VSE++} & 2.4 & 1.48$\pm$0.07 & 0.80$\pm$0.90 & 1.41$\pm$0.74 & 637.10$\pm$20.28 \\
	& \multicolumn{1}{l}{VSRN} & 2.8 & 1.67$\pm$0.07 & 0.14$\pm$0.37 & 1.42$\pm$0.76 & 637.23$\pm$20.35 \\ 
\bottomrule
\end{tabular}
\end{table*}

\subheader{SmoothAP} The observations in Table~\ref{tab:gradcount_smooth} are in line with the observations in Table~\ref{tab:gradcount_triplet} and the evaluation performance in Table~\ref{table:recall}.  
At the optimal convergence point SmoothAP takes approximately one triplet into account when computing the gradient w.r.t.\  $\vec{q}$, which results in close-to or similar performances as the Triplet loss SH.
We also observe the following:  the only experiment where the Triplet loss SH outperforms SmoothAP with a large margin (Table~\ref{table:recall}, row 1.5 and 1.8), is also the experiment where the SmoothAP function takes the highest number of negatives into account when computing the gradient w.r.t. $\vec{q}$ (Table~\ref{tab:gradcount_smooth}, row 1.8). This supports the general observation that the more samples that contribute to the gradient, the lower the final evaluation score.

For the t2i task, we also see that  $C_{SmoothAP}^{0}$ is almost as big as the number of samples (640=$(k=5) \times( |\mathcal{B}| =128)$) in the candidate set, for both datasets and methods.  
Hence, barely any query has a gradient value anymore at the optimal convergence point. 
However, this is not the case for the \ac{i2t} task. 
We conclude that optimizing a ranking metric (i.e.,  AP) with only one positive candidate (as is the case for the \ac{t2i} task), might be too easy to optimize and could result in over-fitting.
Therefore, it is not useful to optimize a ranking task like \ac{ICR} with a ranking-based loss function when there is only one positive candidate per query, which is the case for the \ac{i2t} task.
For the \ac{i2t} task, however, there are barely any queries without a gradient value; here we have $k$ positive candidates per query.

\subheader{Upshot}
In summary, 
\begin{enumerate*}[label=(\arabic*)]
	\item it is important to focus on only one (or a limited) number of (hard) negatives per query during the entire training for the gradient computation, so as to prevent the gradient from being dominated by non-informative or easy negative samples.
	\item  Weighting each negative candidate by its score (as is done in NT-Xent) as opposed to weighting all negative equally (as is done in the Triplet loss) can beneficial for the gradient computation and therefore for the final evaluation score. However, this weighted sum of negatives does not result in the fact that the NT-Xent loss outperforms the Triplet loss SH, which implies that the gradient computation for the NT-Xent is still based on too many non-informative samples.
\end{enumerate*} 


\shrink
\section{Conclusion}
\shrink
We have examined three loss functions from the metric learning field to question if the promising results obtained in metric learning generalize to the \acf{ICR} task.
In contrast with the findings from metric learning, we find that the Triplet loss with semi-hard negative mining still outperforms the NT-Xent and SmoothAP loss.
Hence, the Triplet loss should still be the de facto choice as a loss function for \ac{ICR}; results from metric learning do not generalize directly to \ac{ICR}.
To gain a better understanding of why a loss function results in better performance than others, we have introduced the notion of \AnalysisMethodLong~(\AnalysisMethod).
We have shown that the best performing loss function only focuses on one (hard) negative sample when computing the gradient w.r.t.\ the query and therefore results in the most informative gradient. 
\AnalysisMethod{} suggests that the underperforming loss functions take too many (non-informative) negatives into account, and therefore converge to a sub-optimal point.

The definition of \AnalysisMethod{} uses a threshold value. 
The idea that a candidate contributes to the gradient if its weight value is above a certain threshold is insightful but does not provide the complete picture of how strong the influence of this sample is.
We encourage two directions for future work: 
\begin{enumerate*}[label=(\arabic*)]
	\item Work on more sophisticated methods to determine the influence of (the number of) samples on the gradient w.r.t.\ a query. 
	\item Design new loss functions for the \ac{ICR} task by taking the lessons from \AnalysisMethod{} into account, i.e., loss functions that only take one, or a limited number of, hard negative(s) into account.
\end{enumerate*}
Additionally, we  want to investigate if our findings generalize to fields such as \acf{DPR}~\cite{karpukhin-etal-2020-dense}. \ac{DPR} methods are also mainly optimized by using two data encoders \cite{khattab2020colbert, karpukhin-etal-2020-dense}, for the query and for documents, and the main learning objective is contrasting positive and negative candidates with a query \cite{Gao-2021-rethink, khattab2020colbert, karpukhin-etal-2020-dense, zhan2021dense, chen2021Simplified, formal2021splade}, similar to \ac{ICR}. 

\header{Acknowledgements}
We thank Gabriel Benedict, Mariya Hendriksen, Maria Heuss, Sarah Ibrahimi, and Ana Lucic for feedback and discussions.
This research was supported by the Nationale Politie and the Hybrid Intelligence Center
through the Netherlands Organisation for Scientific Research.
All content represents the opinion of the authors, which is not necessarily shared or endorsed by their respective employers and/or sponsors.

\bibliographystyle{spbasic}
\bibliography{main}

\clearpage

\end{document}